\pdfoutput=1
\documentclass[11pt]{article}

\usepackage[final]{acl}

\usepackage{times}
\usepackage{latexsym}

\usepackage[T1]{fontenc}
\usepackage[utf8]{inputenc}

\usepackage{microtype}

\usepackage{inconsolata}

\usepackage{graphicx}

\usepackage{amssymb}
\usepackage{amsmath}
\usepackage{amsthm}
\usepackage{bbm}
\allowdisplaybreaks

\newtheorem{theorem}{Theorem}
\newtheorem{lemma}{Lemma}
\theoremstyle{definition}
\newtheorem{definition}{Definition}

\usepackage{algorithm}
\usepackage[noend]{algpseudocode}

\usepackage{booktabs}
\usepackage{multicol}
\usepackage{multirow}
\usepackage{graphicx}
\usepackage{textcomp}

\usepackage{svg}
\usepackage{subfig}
\usepackage{pdfpages}

\usepackage{gb4e}
\noautomath % Prevents gb4e from interfering with math symbols

\usepackage{color-edits}
\addauthor{lt}{cyan}
\addauthor{gn}{magenta}
\addauthor{sh}{violet}

\title{What Goes Into a LM Acceptability Judgment? \\ Rethinking the Impact of Frequency and Length}

\author{
 \textbf{Lindia Tjuatja\textsuperscript{1}},
 \textbf{Graham Neubig\textsuperscript{1}},
 \textbf{Tal Linzen\textsuperscript{2}},
 \textbf{Sophie Hao\textsuperscript{2}}
\\
\\
 \textsuperscript{1}Carnegie Mellon University,
 \textsuperscript{2}New York University
\\
 \small{
   \textbf{Correspondence:} \href{mailto:email@domain}{lindiat@andrew.cmu.edu}
 }
}

\begin{document}
\maketitle

    \begin{abstract}
        When comparing the linguistic capabilities of language models (LMs) with humans using LM probabilities, factors such as the length of the sequence and the unigram frequency of lexical items have a significant effect on LM probabilities in ways that humans are largely robust to. Prior works in comparing LM and human acceptability judgments treat these effects uniformly across models, making a strong assumption that models require the same degree of adjustment to control for length and unigram frequency effects. We propose \textit{MORCELA}, a new linking theory between LM scores and acceptability judgments where the optimal level of adjustment for these effects is estimated from data via learned parameters for length and unigram frequency. We first show that MORCELA outperforms a commonly used linking theory for acceptability---SLOR \cite{pauls-klein-2012-large, lauGrammaticalityAcceptabilityProbability2017}---across two families of transformer LMs (Pythia and OPT). Furthermore, we demonstrate that the assumed degrees of adjustment in SLOR for length and unigram frequency overcorrect for these confounds, and that larger models require a lower relative degree of adjustment for unigram frequency, though a significant amount of adjustment is still necessary for all models. Finally, our subsequent analysis shows that larger LMs' lower susceptibility to frequency effects can be explained by an ability to better predict rarer words in context.% 
         \footnote{Our code is available at \url{https://github.com/lindiatjuatja/morcela}.}
    \end{abstract}

\section{Introduction}
{\emph{Are the probabilities provided by language models (LMs) compatible with theories of linguistics and human language processing?}
This is a fundamental question that has implications in fields from psycholinguistics to natural language processing applications,}
and requires understanding of how to relate LM probabilities with quantities associated with human language processing.
{In this work, we consider the relationship between LM probabilities and human judgments of} 
\textit{linguistic acceptability}, and investigate how LM probabilities should be treated when comparing them to human acceptability judgments.

\begin{figure}
    \centering 
    % \small
    \begin{exe}
        \ex \label{ex:good} {\textbf{Acceptable} (Score: 1.19)\\It is silly for one to sing in the shower.}
        \ex \label{ex:middle} {\textbf{Borderline} (Score: 0.00)\\Tanya danced with as handsome a boy as her father.}
        \ex \label{ex:bad} {\textbf{Unacceptable} (Score: \textminus 1.11)\\It seems a cat to be in the tree.}
    \end{exe}
    \caption{English sentences with linguistic acceptability scores reported by \citet{sprouse2013comparison}. Participants where asked to rate sentences on a scale from 1 (least acceptable) to 7 (most acceptable), whose scores were then normalized by participant to a mean of 0 and variance of 1. Scores shown are averaged across participants.}
    \label{fig:acceptability}
\end{figure}

\textit{Acceptability judgments} are speakers' reported perceptions about the well-formedness of utterances, which are often elicited by asking questions such as ``How natural/acceptable/grammatical is this utterance?'' \cite{SprouseBibAcceptabilityJudgments}.
These judgments are typically reported in binary or numerical form \citep{SORACE20051497, Sprouse2007ContinuousAC, sprouse2015three, lauGrammaticalityAcceptabilityProbability2017}, and they are collected through a variety of annotation tasks, such as binary classification, Likert scale scoring, or ranking \cite{schutze2016empirical, sprouse2013comparison}. 
Examples of acceptability judgments, from \citet{sprouse2013comparison}, are provided in \autoref{fig:acceptability}. 
Judgments such as these play a central role in linguistics, where they are used to motivate and evaluate theories of natural language syntax \citep{chomskySyntacticStructures1957}. 

In order to relate LM probabilities with any human behavioral measure, we need a \textit{linking theory} between them to make the two quantities comparable.
Although the existence of a relationship between probability and acceptability has been subject to debate \citep{quineWordObject1960,chomskyQuinesEmpiricalAssumptions1969,pereiraFormalGrammarInformation2000,norvigChomskyTwoCultures2017}, an influential proposal by \citet{lauGrammaticalityAcceptabilityProbability2017} hypothesizes that sentence-level LM probabilities largely reflect linguistic acceptability, but are influenced by word frequency and sentence length in ways that humans are largely robust to. Thus, a linking theory between LM probabilities and human acceptability scores should somehow control for these factors.
Out of the various functions they used to control for length and frequency, \citet{lauGrammaticalityAcceptabilityProbability2017} find that the \textit{syntactic log-odds ratio} (SLOR, \citealp{pauls-klein-2012-large}) served as the best linking theory between acceptability and probabilities from $n$-gram LMs and simple recurrent LMs \citep{elmanFindingStructureTime1990}. 
SLOR controls for unigram frequency and length in a uniform manner across LMs by dividing the probability of the sentence under the LM by the joint unigram frequency, then averaging over all tokens to control for length. However, it is not clear \textit{a priori} that these model-agnostic transformations are the appropriate ones to link LM probabilities and acceptability judgments, nor that these transformations should be held constant across different LMs.

In this work, we first show that the model-agnostic transformations in SLOR may severely underestimate
LM probability correlations with human acceptability judgments.
We propose a new linking theory, \textit{Magnitude-Optimized Regression for Controlling Effects on Linguistic Acceptability} (MORCELA), 
a parameterized linking theory where the effect sizes of length and unigram frequency are automatically estimated from human acceptability judgment data. 
Our experiments first show that MORCELA significantly outperforms SLOR in predicting human acceptability judgments from probabilities calculated by Transformer LMs \cite{vaswaniAttentionAllYou2017} from the Pythia \citep{biderman2023pythia} and OPT \citep{zhang2022opt} families. Our results show a relationship with scale, where larger models exhibit greater correlation with human judgments compared to smaller models in the same family, using the same linking theory. Examining the estimated optimal parameter values of MORCELA reveals that larger models are more robust to length and unigram frequency effects, and thus their probabilities require a lower degree of adjustment when comparing them to human acceptability judgments. We show in particular that larger models' lower reliance on unigram frequency is driven by their improved ability to predict rare words given appropriate context. These results demonstrate that when comparing probability-based LM acceptability judgments to those of humans, controls for factors like length and unigram frequency should be made on a per-model basis.

\section{MORCELA: Acceptability Judgments from LM Probabilities}
To evaluate LMs according to their ability to predict human judgments of linguistic acceptability, we need a \textit{linking function} that takes as input the probability of the sentence under a LM and outputs an acceptability score, which we then correlate with human judgments. This linking function should account for the effects of length and unigram frequency, as noted by \citet{lauGrammaticalityAcceptabilityProbability2017}, which impact LM probabilities in predictable ways that may cause them to deviate from human judgments. Specifically, longer sentences will be assigned a lower probability than to any strictly smaller prefix of the sentence and a sentence containing a rare token will likely have a lower probability compared to one containing a more frequent one, all else being equal. 
% {produces acceptability predictions from LM probabilities in conjunction with length and unigram frequency. Following \citet{lauGrammaticalityAcceptabilityProbability2017}, the linking function should predict that more acceptable sentences, shorter sentences, and sentences with more frequent tokens should have higher LM probabilities}, all else being equal. 

We propose MORCELA (\textbf{M}agnitude-\textbf{O}ptimized \textbf{R}egression for \textbf{C}ontrolling \textbf{E}ffects on \textbf{L}inguistic \textbf{A}cceptability), a parameterized linking function given by 
\begin{equation}
\text{acceptability} \propto \frac{p - \beta u + \gamma}{\ell} \label{eqn:morcela}
\end{equation}
where $\ell$ is the length of a sentence, $p$ is the sentence's LM log probability, $u$ is the sentence's unigram log probability, and $\beta$ and $\gamma$ are learnable parameters. The values of $\beta$ and $\gamma$ can be estimated from human acceptability judgment data by fitting a linear regression model
\[
\text{acceptability} \approx a\frac{p}{\ell} + b\frac{u}{\ell} + c \frac{1}{\ell} + d
\]
and taking $\beta = -b/a$ and $\gamma = c/a$. MORCELA improves upon the \textit{syntactic log-odds ratio} (SLOR, \citealp{pauls-klein-2012-large}), widely used as a linking function for predicting acceptability judgments \citep{lauGrammaticalityAcceptabilityProbability2017,lauAcceptabilityInContext, sprouseColorlessGreenIdeas2018, kann-etal-2018-sentence, kumar-etal-2020-iterative, misra2024language, lu-etal-2024-syntactic}, by allowing for arbitrary linear relationships between the variables $p$ and $u$ via the parameters $\beta$ and $\gamma$. We argue here that MORCELA mitigates overcorrections for length and frequency effects from SLOR, and our main experiment shows that MORCELA scores are significantly more correlated with $z$-normalized human acceptability ratings than SLOR scores.

\subsection{Prior Work: SLOR}
SLOR was proposed as a linking function by \citet{lauGrammaticalityAcceptabilityProbability2017}, who compare it against several other linking functions on acceptability judgments from multiple sources. SLOR predicts the acceptability rating of a sentence to be given by 
\begin{equation}
\text{acceptability} \propto \frac{p - u}{\ell} \label{eqn:slor}
\end{equation}
where $p$, $u$, and $\ell$ are defined as above. Intuitively, the SLOR score of a sentence is the average log probability assigned to its tokens, adjusted for frequency. It uses $p$ as an initial estimate of acceptability, and incorporates $\ell$ and $u$ under the assumption that long sentences and sentences with rare words have lower LM probabilities, but not lower human acceptability judgments, than short sentences or sentences without rare words.

\subsection{MORCELA}
The normalizations involved in SLOR are based on specific assumptions about the impact of length and frequency on LM probabilities, namely that LM probabilities and unigram frequencies should have equal importance on the resulting acceptability score, and that taking the geometric mean of each token's probability under the LM largely eliminates the impact of sentence length.
However, it is unclear \textit{a priori} whether these assumptions hold, and furthermore whether they hold uniformly across models. 

MORCELA relaxes these assumptions by allowing for arbitrary linear relationships between LM probabilities and unigram frequencies, expressed via the parameters $\beta$ and $\gamma$. These parameters can be understood as mitigating overcorrections for frequency and length effects by SLOR, respectively. To see this, let us rewrite equations (\ref{eqn:morcela}) and (\ref{eqn:slor}) as follows:
\[
\text{MORCELA} = \text{SLOR} + \underbrace{(1 - \beta) \frac{u}{\ell}}_{\text{frequency}} + \underbrace{\gamma\frac{1}{\ell}}_{\text{length}}
\]
The ``frequency'' term, controlled by $\beta$, adjusts SLOR according to the average unigram probability of the sentence's tokens, while the ``length'' term, controlled by $\gamma$, provides an adjustment to SLOR that is inversely proportional to the sentence's length.

\section{Main Experiment}

How much does optimizing the relative effect of length and unigram frequency via MORCELA impact fit of LM acceptability scores to human judgments?
To investigate this, we correlate the LM acceptability scores from MORCELA to gradient human judgments across LMs of varying sizes and compare the resulting correlation to two baseline linking functions: log probabilities and SLOR.

\subsection{Models}

We evaluate models of varying sizes from the Pythia Scaling Suite \cite{biderman2023pythia} and Open Pre-Trained Transformers (OPT, \citealt{zhang2022opt}) families. Both families of models are decoder-only autoregressive transformer LMs. We test all eight sizes of Pythia models (70M--12B parameters) and all but the two largest OPT models (125M--30B parameters). Models within each family were trained on the same pretraining corpus: Pythia models were trained on The Pile \cite{gao2020pile}, whereas the OPT models were trained on a concatenation of data from subsets of the RoBERTa training corpus \cite{zhuang-etal-2021-robustly}, The Pile \cite{gao2020pile}, and PushShift.io Reddit \cite{baumgartner2020pushshift, roller-etal-2021-recipes}. Both families of models saw $\approx{300}$B tokens during training. We list additional model hyperparameter details in \autoref{sec:params}.

\subsection{Unigram Frequency Estimation}
\label{sec:unigram-frequency-estimation}
As input to the various linking functions, we need to calculate the LM probability $p$, the unigram probability $u$, and length $\ell$ of the sentence in tokens.\footnote{Technically, for the OPT models this involves calculating the probability conditioned on only the beginning of sequence (BOS) token. Pythia models were trained without a BOS token, so to calculate $p$ we do not append an additional BOS token to the input sequence, and instead exclude the first token's probability when calculating $p$ and $u$, though subsequent tokens in $p$ are calculated with the first token provided in the context. We also exclude the first token when calculating the length of the sentence $\ell$.} 
To calculate $u$, we need to measure the frequency of tokens as they appear in the training corpus of the LM. This is easily done for the Pythia models as the training corpus (The Pile, \citealt{gao2020pile}) is publicly available. However, since this is not the case for the OPT models, we instead look at text generated from the largest OPT model we test (OPT-30B) as a proxy for the training corpus, with the intuition that the distribution of generated text is largely similar to that of its pretraining corpus. We estimate token unigram frequency by aggregating the probability of a token being generated in each position of a sequence of arbitrary length, then averaging this value over a large number of generated sequences ($n=100000$). 
% Sequences are generated from just the BOS token as context and are set to the length (in tokens) of the longest sentence in the dataset we evaluate on, which in this case was 34 tokens.
We provide additional details for this estimation process in \autoref{sec:unigrahams}.

\subsection{Dataset}
We evaluate acceptability predictions and fit MORCELA parameters using data from \citet{sprouse2013comparison}, which contain acceptability judgments for example sentences from the \textit{Linguistic Inquiry}, a leading theoretical linguistics journal. We use judgments reported on a 1--7 Likert scale by native English speakers in the United States, $z$-normalized by annotator.\footnote{For example, the acceptability score of 1.19 for sentence (1) in \autoref{fig:acceptability} means that this sentence was judged to be 1.19 standard deviations more acceptable than the mean acceptability of sentences in the dataset.} To ensure balance, we limit our dataset to minimal pairs of acceptable and unacceptable sentences that \citet{sprouse2013comparison} have determined to have equal semantic plausibility. After filtering out unpaired sentences as well as sentences with missing data, we obtain a final dataset of acceptability judgments for 1450 sentences.

\subsection{Fitting and Evaluating Linking Functions}
For each linking function we examine, we calculate the correlation (Pearson's $r$) between the LM acceptability scores generated by the function and z-normalized Likert scale human judgments. For functions with learned parameters, we train and evaluate linear regression models using 5-fold cross validation (with shuffling), and report the average correlation over each test fold. To calculate an upper bound for correlation, we randomly split judgments per sentence into two groups, which yields an inter-group correlation of $r = 0.860$.

\section{Results}
\begin{figure*}[h]
    \centering
    {\includegraphics[width=\textwidth]{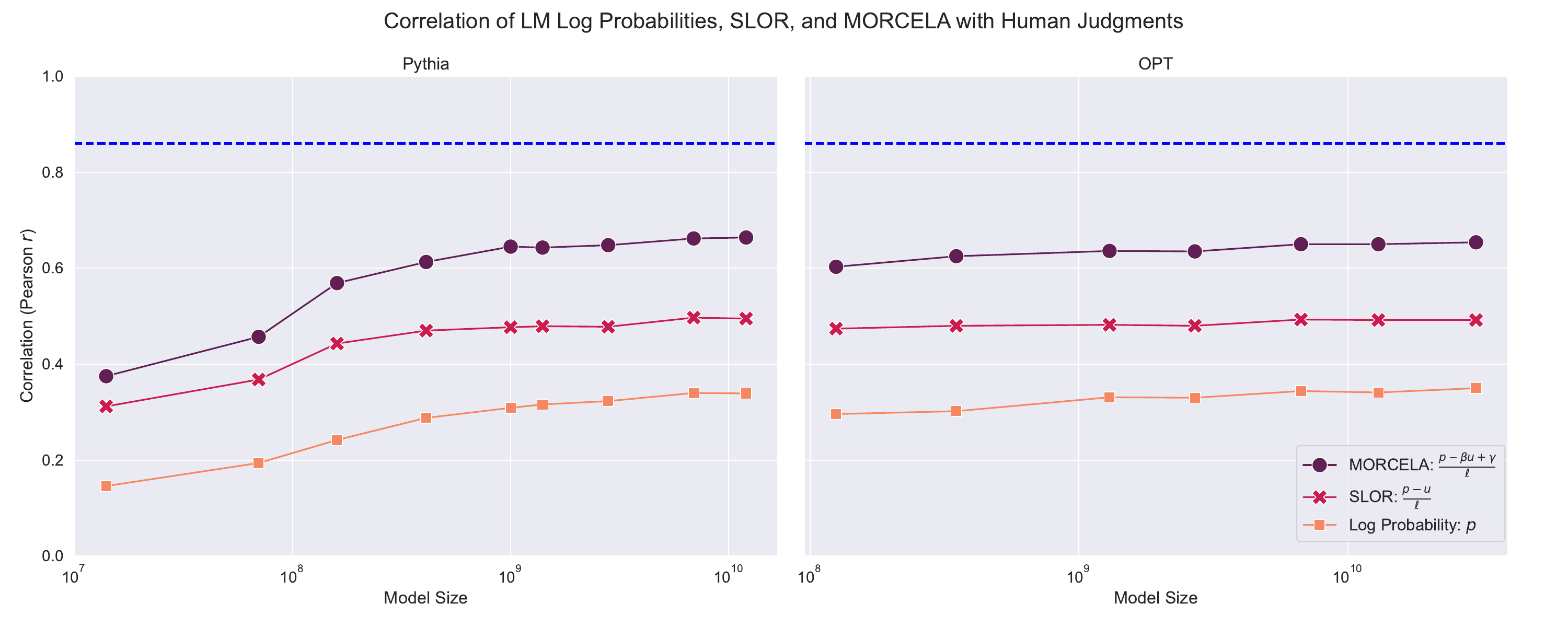}}
    \caption{Correlation of LM acceptability scores with human judgments using raw log probabilities, SLOR, and MORCELA. The blue dashed line indicates inter-group correlation between randomly partitioned participant ratings ($r=0.860$). MORCELA consistently outperforms SLOR, with up to $+\Delta0.17$ gain in correlation.}%
    \label{fig:slor}
\end{figure*}

\begin{figure*}[h]
    \centering
    {\includegraphics[width=\textwidth]{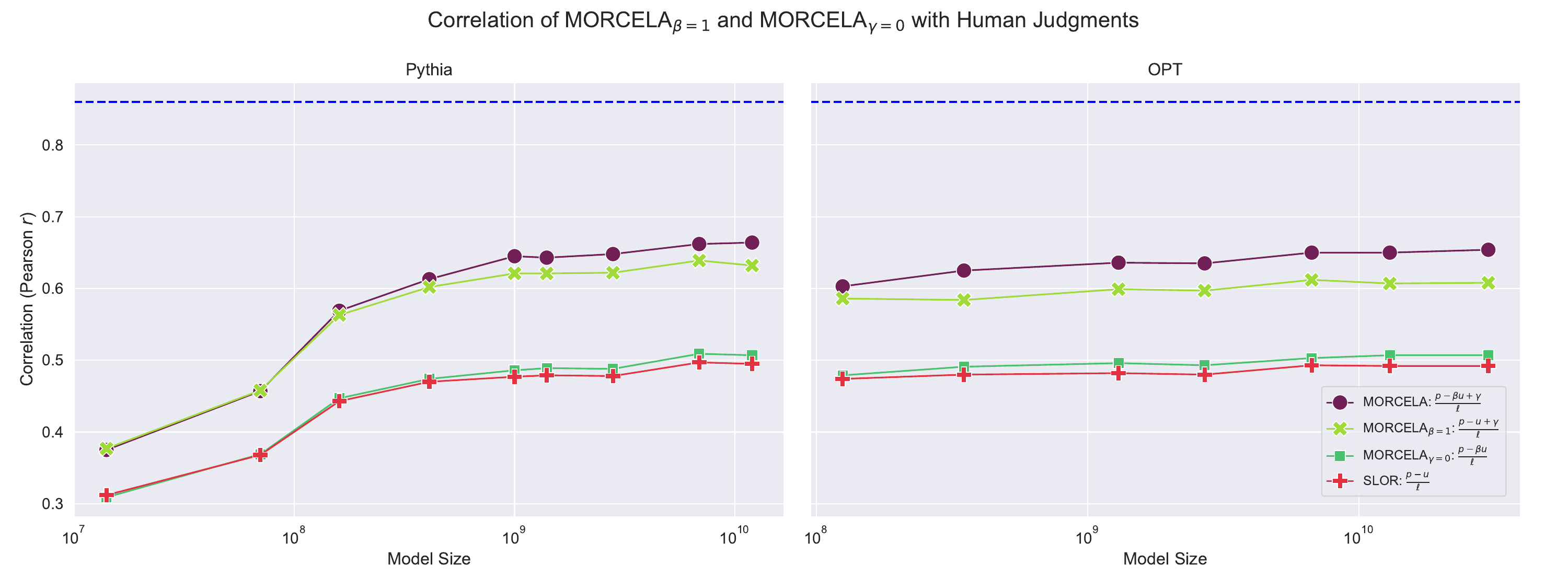}}
    \caption{Comparison of SLOR and MORCELA with linking functions where either length or unigram normalization are set to their default values ($\beta=1$, $\gamma=0$) and the other is allowed to vary. Allowing the length-normalized intercept $\gamma$ to vary on its own (MORCELA$_{\beta = 1}$) leads to similar performance to MORCELA, though models still benefit from additionally varying the unigram coefficient $\beta$.}%
    \label{fig:ablations}
\end{figure*}
We first compare MORCELA to two baseline linking functions (raw log probabilities and SLOR), then assess the impact that parameterizing either unigram frequency or length has on correlation with human judgments. 

\subsection{MORCELA vs. SLOR}
\autoref{fig:slor} shows correlation of acceptability scores using log probabilites, SLOR, and MORCELA across varying sizes of Pythia and OPT models. 
There is a general increasing monotonic trend with size, though to a lesser degree with the OPT models as smaller OPT models have a higher correlation with humans compared to similarly sized Pythia models. Nevertheless, overall trends regarding the relative performances of the different linking functions, and how they change with scale, are similar. 

Across all models, the addition of the two learned parameters in MORCELA leads to a significant gain in correlation with human judgments. We observe up to $+\Delta0.33$ increase from raw log probabilities and $+\Delta0.17$ from SLOR with Pythia-6.9B and 12B, which amounts to a $46\%$ relative error reduction from SLOR with respect to the inter-group correlation upper bound. 
As models get larger (and correspondingly, generally better at predicting human judgments), we also observe greater differences in correlation between SLOR and MORCELA. This suggests that larger models with higher baseline correlation with humans (as demonstrated by higher raw log probability correlation) reap greater benefits from the additional parameterization.

\subsection{Parameter Ablation Study}
The performance gap between SLOR and MORCELA clearly shows that the assumed values ($\beta=1$, $\gamma=0$) in SLOR are non-optimal across all models, and especially so for larger ones.
But how important is the optimization of either parameter in improving fit to human judgments? To answer this, we perform ablations to MORCELA, where either length or unigram normalization are set to their default values ($\beta=1$, $\gamma=0$) and the other is allowed to vary.

The results of these ablations are shown in \autoref{fig:ablations}, where MORCELA$_{\beta=1}$ optimizes the value of the length-normalized intercept $\gamma$ given the default weight for unigram frequency, and vice versa for MORCELA$_{\gamma=0}$. We find that optimizing the unigram coefficient $\beta$ without a length intercept (MORCELA$_{\gamma=0}$) leads to little to no gain in performance. In contrast, adding the length-normalized intercept $\gamma$ while keeping the unigram coefficient $\beta$ fixed (MORCELA$_{\beta=1}$) can---for the smallest Pythia models---reach the performance of MORCELA, though the difference between MORCELA$_{\beta=1}$ and MORCELA tends to grow as models become larger. These results are similar to those from \citet{lauAcceptabilityInContext} which found that a linking function that only depends on log probability and sequence length led to a higher correlation with human judgments compared to SLOR for bidirectional models (e.g. BERT) as well as GPT-2, though their linking function scales length with an exponent to dampen the impact of large values.

Given that we can get a large gain in correlation from optimizing the length intercept $\gamma$ alone, a natural question that follows is whether this gain in correlation is significant enough to warrant the extra degree of freedom that comes with additionally varying $\beta$. Using two model selection criteria---Akaike information criterion (AIC, \citealt{akaike1974new}) and Bayes information criterion (BIC, \citealt{Schwarz1978EstimatingTD}), which take into account both model fit and number of predictors---we find that it is: MORCELA is preferred over MORCELA$_{\beta=1}$ for all but one LM (Pythia-14M).\footnote{We include details for calculation and values of AIC and BIC for each linking function per LM in Appendix \ref{sec:aic_bic}.}
Thus, while the addition of the length-normalized intercept $\gamma$ on its own can significantly correlation with human judgments, adding the unigram coefficient $\beta$ in conjunction with $\gamma$ is still preferred.

\begin{figure}[h]
    \centering
    {\includegraphics[width=1\linewidth]{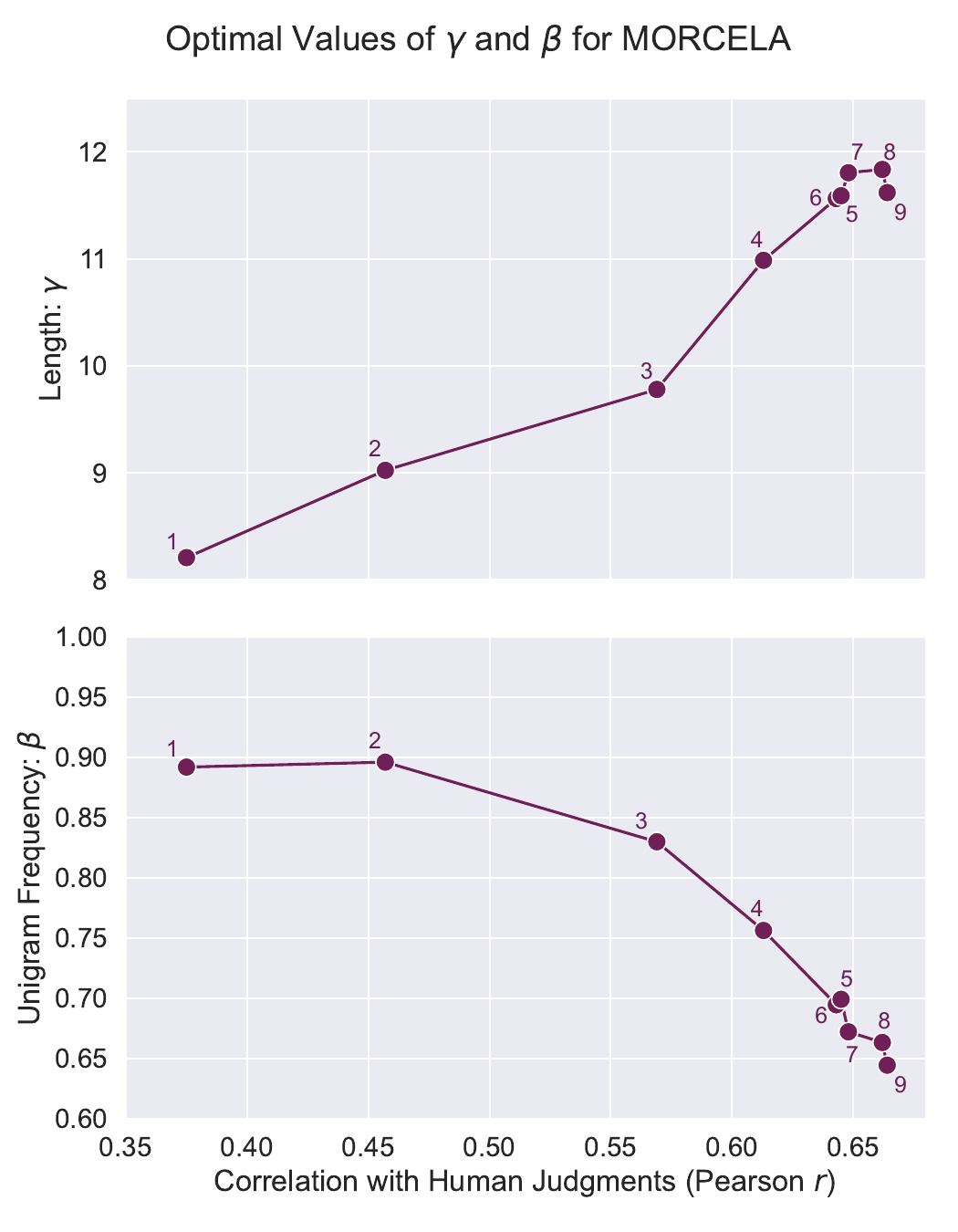}}
    \caption{Optimal values of $\gamma$ (top) and $\beta$ (bottom) versus correlation with human judgments across for MORCELA, fit using all of the data. As models become better correlated with human judgments, $\gamma$ increases and $\beta$ decreases. Points shown are from Pythia models (numbered from smallest to largest), though these trends generally hold for the OPT models as well (see Appendix \ref{sec:more-figs}).}%
    \label{fig:genslor-params}
\end{figure}

\subsection{Trends in Length and Unigram Frequency Effects Across Models}
The above results tell us that the coefficients used by SLOR to control for length and unigram frequency---namely an equal weighting of LM log probabilities and unigram log probabilities and the lack of a length-normalized intercept---are non-optimal, and that the impact of turning these controls into tuned parameters impacts models to varying degrees. However, looking at correlation alone does not tell us about \textit{how} these controls are non-optimal. We inspect the learned optimal values of the unigram coefficient $\beta$ and length-normalized intercept $\gamma$ to see whether the assumed values in SLOR are under- or overestimating the impact of these confounds across models. 

\autoref{fig:genslor-params} shows the optimal values of $\gamma$ and $\beta$ for MORCELA, fit using all the data. All values of $\gamma$ are positive, and grow larger as models become better correlated with human judgments. Qualitatively, the observation that $\gamma$ is positive indicates that naively normalizing by dividing by the length of the sentence is actually \textit{overcorrecting} for length. A larger positive $\gamma$ more dramatically increases acceptability scores of shorter sentences relative to longer ones. This counteracts the division by length, which on its own increases the scores of longer sentences.
% \ltcomment{need more concrete intuition for this}

Like $\gamma$, for the unigram coefficient $\beta$ we see a trend with respect to correlation (and thus, to a large extent, model size), though in this case the value of $\beta$ decreases as correlation increases. Notably, all values of $\beta$ are less than the default value of 1. This, too, shows that the assumed impact of unigram frequency as used in SLOR is an overestimate, and that larger models tend to require less adjustment for unigram frequency. Additionally, we find that trends across both $\gamma$ and $\beta$ also hold among various other linking functions that parameterize $\beta$ and/or $\gamma$, as shown in \autoref{fig:all-params} in Appendix \ref{sec:more-figs}.

\section{Ability to Predict Infrequent Tokens Explains Impact of Unigram Frequency}
\begin{figure*}[!h]
    \centering
    {\includegraphics[width=1\textwidth]{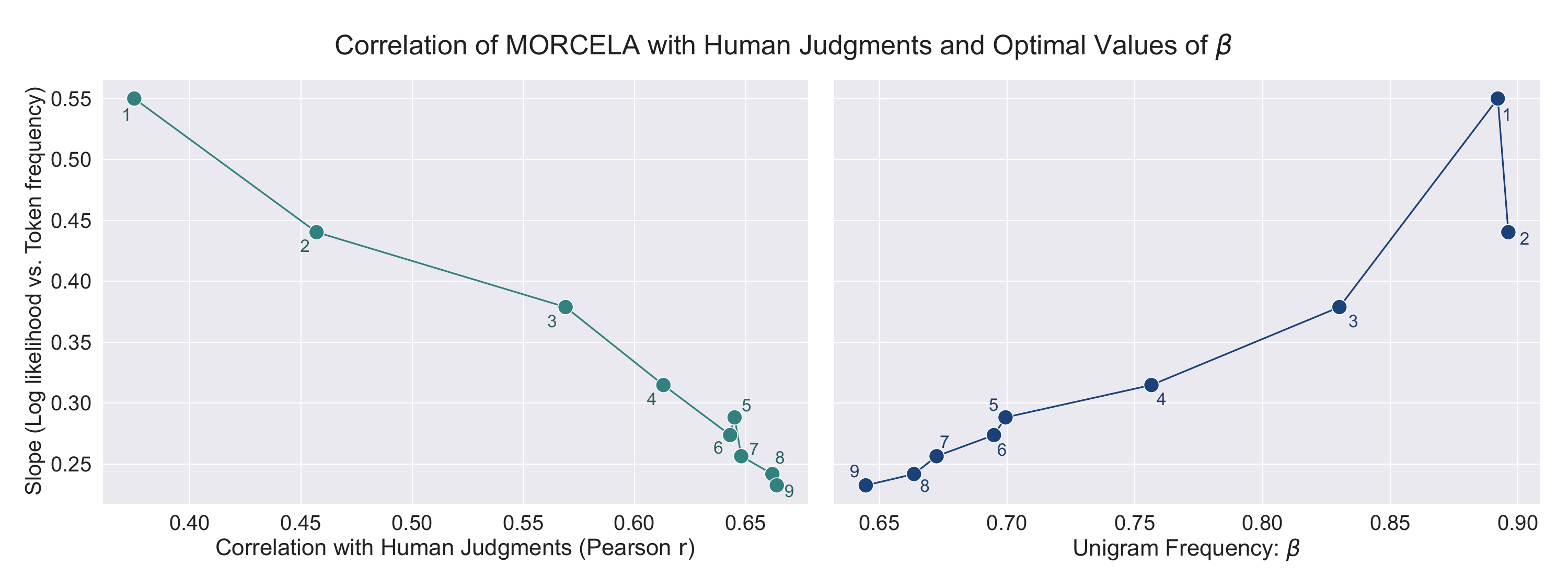}}
    \caption{The slope of LM conditional log likelihood per token vs. token unigram frequency plotted against the correlation (average correlation using 5-fold CV) and optimal unigram coefficient $\beta$ values of MORCELA (fit using all the data). Each point represents a single Pythia model, numbered from smallest to largest. Models that are better at predicting rarer tokens in context should have a \textit{lower} slope value, which we find is correlated with higher correlation with human judgments and generally lower $\beta$.}%
    \label{fig:cross-ent}
\end{figure*}
As we have just shown, as models get larger and better at predicting human acceptability judgments, the smaller the relative importance of unigram frequency becomes. One possible explanation for this is that models that are better predictors of acceptability are so because they are better at predicting more infrequent tokens in context, and as a result are more robust to the effect of unigram frequency. The intuition behind this is that while some tokens may be very rare within the distribution of the entire corpus (e.g. names of chemical compounds), they may be relatively frequent given a specific context (e.g. within a scientific article). Thus, if a model is better able to predict such cases of tokens by utilizing context, they should no longer need to be controlled as heavily for unigram frequency.

To test this hypothesis, we first need a way to quantify the ability of a LM to predict rarer tokens in context. We operationalize this by correlating the LM's conditional log likelihood over instances of tokens with the unigram log-probability of those tokens. As our corpus to calculate conditional log likelihood over, we use a portion ($\sim100$ million tokens) of the test set of The Pile \cite{gao2020pile}, the training corpus of the Pythia models. For unigram log-probabilities, we use counts from the entire training split (as in our calculations of $p_{U}(S)$). We do this for all sizes of models in the Pythia suite. To calculate conditional log likelihood, we use a sliding window of the max sequence length of the Pythia models (2048 tokens) with a stride of 1024. As before, since Pythia was trained without a BOS token, the log likelihood of the first token in a document is not considered. 

If our hypothesis---that LMs better at predicting rarer tokens in context are more robust to unigram frequency effects---holds, we would expect that models that are worse at predicting human acceptability judgments have \textit{lower} conditional log likelihood for more infrequent tokens compared to models that are better fits to judgments, and vice versa. In other words, if we were to plot conditional log probability versus unigram log-probability, we should see that models with higher correlation with human judgments and lower values of $\beta$ have a steeper positive slope.

We find that this prediction largely holds, as shown in \autoref{fig:cross-ent}. In general, as models get better at predicting rarer tokens in context, i.e. a higher log likelihood on aggregate for rarer tokens and thus a less positive slope between log likelihood and token unigram frequency, they show greater correlation with human judgments as well as increasingly lower values of $\beta$.

\section{Discussion}
Our method MORCELA lies within the context of a large body of work that examines how well (neural) language models reflect human-like language processing. We now discuss our work's relation to previous studies evaluating LMs as psycholinguistic subjects and the application of parameterized linking theories in this setting, and compare our results to related findings from comparisons of LM surprisal to reading times. 

\subsection{Methods for Evaluating LMs as Psycholinguistic Subjects}
A common methodological setup in comparing the linguistic capabilities of language models to humans is the targeted syntactic evaluation paradigm \citep[inter alia]{linzen-etal-2016-assessing, marvin-linzen-2018-targeted, gulordava-etal-2018-colorless}. In this setup, probability-based LM judgments are considered consistent with those of humans if they assign a higher probability to acceptable sentences compared to their minimally different unacceptable counterparts. Datasets such as BLiMP \cite{warstadtBLiMPBenchmarkLinguistic2020}, as well as evaluation frameworks like SyntaxGym \cite{gauthierSyntaxGymOnlinePlatform2020}, follow this paradigm. A notable feature of many of these works is the use of a forced choice, binary judgment setup, which comes with the assumption that one sentence is more acceptable than the other. Thus, in these evaluations the information about relative differences within and across pairs is not present, though the use of minimal pairs in itself does not require this. For example, work by \citet{leong-linzen-2023-language} correlate the difference in LM probabilities between the acceptable and unacceptable sentences within a minimal pair with the difference in gradient human judgments. Nevertheless, the use of minimal pairs makes strictly controlling for length and frequency effects possible at the data construction stage.

Our experimental setting is most similar to work such as \citet{lauGrammaticalityAcceptabilityProbability2017}, \citet{lauAcceptabilityInContext}, \citet{javier-vazquez-martinez-etal-2023-evaluating}, \citet{misra2024language}, and \citet{lu-etal-2024-syntactic}, which instead correlate (transformed) LM probabilities directly to gradient human judgments on individual sentences. However, while this setting more easily allows for greater granularity in judgments across a wider range of examples, it requires a linking function that either assumes or estimates the effects of various factors on LM probabilities. Commonly used linking functions in these settings, such as SLOR, bake in assumptions about length and frequency effects, namely that length can be controlled for by dividing by the number of tokens in the sequence, and that unigram frequency and LM probabilities should have equal weight.
Our work challenges these assumptions further by instead estimating this effect directly from acceptability data via a parameterized linking function. 

\subsection{Linking Functions}
More generally, a linking function between measured quantities from LMs and humans is a way to control for asymmetries in effects of factors external to the construct of interest. 
In the case of LM probabilities and acceptability judgments, we expect that LM probabilities over sentences are impacted by unigram frequency and the length in ways that humans are thought to be largely robust to \cite{lauGrammaticalityAcceptabilityProbability2017, Goodall_2021}.

We can draw parallels to methods in comparing LM word-level surprisal with measures of incremental sentence processing (e.g. eye tracking, reading and reaction times), where instead there are external effects on the human side, such as the length or predicatbility (estimated using a statistical language model) of a word in reading time experiments \cite{SMITH2013302, goodkindPredictivePowerWord2018, wilcox2021targeted, meister-etal-2021-revisiting}. However, unlike other works comparing LMs to human acceptability judgments that assume the strength and quality of effects (specifically of length and unigram frequency), it is standard for the parameters associated with the covariates in these studies are learned and fit per participant. MORCELA can be viewed as following a similar methodology, where we instead fit parameters per model to correct for model-side effects. 

Nevertheless, MORCELA, like SLOR, still makes the assumption that the form of the relationship between LM probabilities and acceptability is log-linear. \citet{meister-etal-2021-revisiting} find evidence for a super-logarithmic relationship between LM probabilities and reading times, as well as binary acceptability judgments, differing from our gradient judgment setting; future work could explore other forms to fit between probabilities and gradient judgments.

\subsection{Impact of Scale on Similarity with Humans}
MORCELA demonstrates that the strength of length and unigram frequency effects (1) are not uniform across models, (2) are overestimated by the default values in SLOR, and (3) show a trend with scale. 
As models become larger and generally better predictors of human acceptability judgments (up to a certain point), the less they need to be controlled for unigram frequency effects. 
In contrast, prior work by \citet{ohschuler23tacl} observe the opposite trend with respect to scale when comparing LM surprisal with reading times, with larger models serving as poorer fits to humans. In follow-up work they found that this trend can be explained by frequency, with the inverse correlation between model size and reading times being the strongest amongst the least frequent words \cite{ohetal24eacl}. Similar to our analysis, they show that this is driven by the ability of larger LMs to more accurately predict rare words.

We hypothesize that the seeming paradox between more human-like judgments vs. less human-like reading time predictions may be a consequence of the role of predictability in offline and online language processing. In the case of reading times, it may be that frequency effects at the word level are important for humans, and thus models can be “too good” at predicting rare words relative to humans, whereas this is may not the case---at least, to the same extent---in predicting acceptability judgments.

\section{Conclusion}
In this work, we reexamine the assumptions made by commonly used linking theories such as SLOR in evaluating LMs' fit to human acceptability judgments. We introduce a new linking theory, MORCELA, which parameterizes controls for length and unigram frequency, and learns the optimal values for these controls from acceptability data via linear regression. By adding two simple, interpretable parameters, MORCELA drastically improves correlation with human judgments compared to SLOR, showing that SLOR greatly underestimates correlation between LM and human acceptability scores. An inspection of the optimal values of these parameters shows that the magnitude of correction for confounds in SLOR overestimate the impact of frequency and length in Transformer LMs, and that this overestimation is greater as models grow larger. Finally, we show that LMs' robustness to unigram frequency effects can be explained by their ability to predict rarer words in context. 

Our findings suggest that evaluations of probability-based LM acceptability judgments should account for model-specific qualities with respect to factors like frequency and length, and that doing so reveals that LMs may be better correlated with human judgments than previously thought. However, there is still a sizable gap between the maximum correlation between LMs and human judgments and the correlation between annotators. Future work could investigate what additional factors/transformations could lead to closer correspondence between LMs and humans, and further integrate these insights into training more cognitively plausible models.

\section{Limitations}
Our evaluations are limited to two model families trained on predominantly English data on judgments of English sentences by English AMT workers with a US-based location \cite{sprouse2013comparison}. Thus, there is no guarantee that our results would hold for models or data in other languages/in a multilingual setting. The size of our data ($n=1450$) is relatively small, so while we expect general trends with respect to scale to hold, the actual values of the optimized parameters may change with larger and more varied data.

\section{Ethics Statement}
This work uses publicly available models and data and does not release any new artifacts. For the human acceptability judgment data, we point readers to \citet{sprouse2013comparison} for details on data collection. We do not foresee any negative ethical consequences for our work. 

\section*{Acknowledgments}
This material is based upon work supported by the National Science Foundation (NSF) under Grant No. BCS-2114505 and a gift from Amazon AWS. This work was also supported in part through the NYU IT High Performance Computing resources, services, and staff expertise. We would also like to thank Kanishka Misra, members of NeuLab and Caplab, and our reviewers for their helpful feedback on our work.

% Bibliography entries for the entire Anthology, followed by custom entries
%\bibliography{anthology,custom}
% Custom bibliography entries only
\bibliography{custom}

\appendix

\section{Model Hyperparameters}
\label{sec:params}
Hyperparameters of the Pythia and OPT models examined in this work are shown in \autoref{tab:params}.
\begin{table}[ht!]
    \centering
    \small
    \begin{tabular}{lrrrr} \toprule
    Model Variant & \#L & \#H & $d_{\text{model}}$ & \#Parameters \\ \midrule
    OPT 125M & 12 & 12 & 768 & $\sim$125M \\
    OPT 350M & 24 & 16 & 1024 & $\sim$350M \\
    OPT 1.3B & 24 & 32 & 2048 & $\sim$1.3B \\
    OPT 2.7B & 32 & 32 & 2560 & $\sim$2.7B \\
    OPT 6.7B & 32 & 32 & 4096 & $\sim$6.7B \\
    OPT 13B & 40 & 40 & 5120 & $\sim$13B \\
    OPT 30B & 48 & 56 & 7168 & $\sim$30B \\ \midrule
    Pythia 14M & 6 & 4 & 512 & $\sim$14M \\
    Pythia 70M & 6 & 8 & 512 & $\sim$70M \\
    Pythia 160M & 12 & 12 & 768 & $\sim$160M \\
    Pythia 410M & 24 & 16 & 1024 & $\sim$410M \\
    Pythia 1B & 16 & 8 & 2048 & $\sim$1B \\
    Pythia 1.4B & 24 & 16 & 2048 & $\sim$1.4B \\
    Pythia 2.8B & 32 & 32 & 2560 & $\sim$2.8B \\
    Pythia 6.9B & 32 & 32 & 4096 & $\sim$6.9B \\
    Pythia 12B & 36 & 40 & 5120 & $\sim$12B \\ \bottomrule
    \end{tabular}
    \caption{Hyperparameters of model variants examined in this work. \#L, \#H, and $d_{\text{model}}$ respectively refer to number of layers, number of attention heads per layer, and embedding size.}
    \label{tab:params}
\end{table}

\section{Unigrahams Estimator}
\label{sec:unigrahams}

% n = 1M, l = 34

\algrenewcommand\algorithmicloop{\textbf{repeat}}
\begin{algorithm}
    \caption{Unigrahams Estimator}\label{alg:unigrahams}
    \textbf{Inputs:} LM $m$, prompt $x$, response length $\ell$ \\
    \textbf{Parameters:} Number of samples $n$, vocabulary $\mathbb{V}$ \\
    \textbf{Output:} Frequency estimates $f_\ell(\cdot \mathrel{|} x):\mathbb{V} \to \mathbb{R}$
    \begin{algorithmic}
        \ForAll{$w \in \mathbb{V}$}
            \State $f_\ell(w \mathrel{|} x) \gets 0$
        \EndFor
        \Loop\ $n$ \textbf{times}
            \State $y \gets x$
            \Repeat
                \ForAll{$w \in \mathbb{V}$}
                    \State $f_\ell(w \mathrel{|} x) \gets f_\ell(w \mathrel{|} x) + \frac{\mathbb{P}_m[w \mathrel{|} y]}{n}$
                \EndFor
                \State Sample a token $v \sim \mathbb{P}_m[\cdot \mathrel{|} y]$
                \State $y \gets yv$
            \Until{$|y| = \ell + |x|$}
        \EndLoop
        \Return $f_\ell(\cdot \mathrel{|} x)$
    \end{algorithmic}
\end{algorithm}

This section proposes a novel technique, the \textit{unigrahams estimator}, for estimating the unigram distribution of an LM without access to its training corpus. The unigrahams estimator estimates the unigram distribution for text generated by an LM of a given length $\ell$, which we assume approximates the unigram distribution of the LM's training corpus. Given a prompt $x$, we generate a number $n$ of responses to $x$, counting the number of occurrences of each token in each response. These frequency counts are weighted by LM probabilities in the following sense: if, during the generation process, the LM assigns a next-token probability of $q$ to token $w$, then we assume that $q$ instances of $w$ have occurred in this position of the generated text. A full description of the unigrahams estimator is given in \autoref{alg:unigrahams}, and its correctness is proven in \autoref{sec:unigrams-proof}.

As mentioned in \autoref{sec:unigram-frequency-estimation}, we use the unigrahams estimator to estimate the unigram distribution of the OPT training corpus from the OPT-30B model. We use parameter values of $n = 10^6$ and $\ell = 34$, the latter being the length of the longest sentence in our dataset of acceptability judgments from \citet{sprouse2013comparison}.

\subsection{Theoretical Analysis}
\label{sec:unigrams-proof}

We now prove the correctness of the unigrahams estimator. Let $m$ be a LM, and let $\mathbb{P}_m[y \mathrel{|} x]$ denote the probability that $m$ will generate response $y$ to prompt $x$. Let $|y|$ denote the length of $y$ (in tokens), let $y_{:i}$ denote the first $i$ tokens of $y$, and for each token $w$ in vocabulary $\mathbb{V}$, let $|y|_w$ denote the number of occurrences of $w$ in $y$. Below we give a formal definition of a LM's unigram distribution.

\begin{definition}
    The \textit{length-$\ell$ unigram frequency} of a token $w \in \mathbb{V}$ with respect to prompt $x$ and LM $m$ is the expected number of times $w$ occurs in responses to $x$ of length $\ell$:
    \[
    f_\ell(w \mathrel{|} x) := \mathbb{E}_{|y| = \ell}[|y|_w]
    \]
    where $y \sim \mathbb{P}_m[\cdot \mathrel{|} x]$.
\end{definition}

Our goal is to show that the unigrahams estimator is an unbiased estimator of $f_\ell(w \mathrel{|} x)$ for each $w \in \mathbb{V}$. This result is stated as follows.
\begin{theorem}
    For all $\ell \geq 1$ and $w \in \mathbb{V}$,
    \[
    f_\ell(w \mathrel{|} x) = \mathbb{E}_{|y| = \ell - 1} \left[ \sum_{i = 0}^{\ell - 1} \mathbb{P}_m[w \mathrel{|} xy_{:i}] \right]
    \] \label{thm:unigrahams}
\end{theorem}
To see why \autoref{thm:unigrahams} is the desired result, observe that \autoref{alg:unigrahams} returns the mean of
\[
\sum_{i = 0}^{\ell - 1} \mathbb{P}_m[w \mathrel{|} xy_{:i}]
\]
for a sample of $y$s of length $\ell - 1$ drawn from $\mathbb{P}_m[\cdot \mathrel{|} x]$. This is an unbiased estimator of the right-hand side of \autoref{thm:unigrahams}, whence it follows that proving \autoref{thm:unigrahams} suffices for verifying the correctness of the unigrahams estimator.

To that end, we note that \autoref{thm:unigrahams} is straightforwardly derived from the following lemma.

\begin{lemma}
    For all $\ell > 1$ and $w \in \mathbb{V}$,
    \[
    f_{\ell}(w \mathrel{|} x) = f_{\ell - 1}(w \mathrel{|} x) + \mathbb{E}_{|y| = \ell - 1}[\mathbb{P}_m[w \mathrel{|} xy]]
    \] \label{lemma}
\end{lemma}

\begin{proof}[Proof of \autoref{lemma}]
    Observe:
    \begin{align*}
        &\mathrel{\phantom{=}} f_\ell(w \mathrel{|} x) \\
        &= \sum_{|y| = \ell} |y|_w\mathbb{P}_m[y \mathrel{|} x] \\
        &= \sum_{|y| = \ell - 1} \left[ (|y|_w + 1)\mathbb{P}_m[yw \mathrel{|} x] \mathrel{+} \vphantom{\sum_v} \right. \\
        &\mathrel{\phantom{=}} \phantom{\sum_{|y| = \ell - 1}} \left.\hspace{2em} |y|_w\sum_{v \in \mathbb{V} \backslash \lbrace w \rbrace} \mathbb{P}_m[yv \mathrel{|} x] \right] \\
        &= \sum_{|y| = \ell - 1} (|y|_w + \mathbb{P}_m[w \mathrel{|} xy])\mathbb{P}_m[y \mathrel{|} x] \\
        &= \mathbb{E}_{|y| = \ell - 1}[|y|_w + \mathbb{P}_m[w \mathrel{|} xy]] \\
        &= f_\ell(w \mathrel{|} x) + \mathbb{E}_{|y| = \ell - 1}[\mathbb{P}_m[w \mathrel{|} xy]]\text{.}
    \end{align*}
\end{proof}

\begin{proof}[Proof of \autoref{thm:unigrahams}]
    We induct on $\ell$. To prove the $\ell = 1$ case, we simply observe that
    \[
    f_1(w \mathrel{|} x) = \mathbb{P}_m[w \mathrel{|} x]\text{.}
    \]
    Now suppose \autoref{thm:unigrahams} holds for some value of $\ell$. Observe that
    \begin{align*}
    &\mathrel{\phantom{=}} \mathbb{E}_{|y| = \ell - 1} \left[ \sum_{i = 0}^{\ell - 1} \mathbb{P}_m[w \mathrel{|} xy_{:i}] \right] \\
    &= \mathbb{E}_{|y| = \ell} \left[ \sum_{i = 0}^{\ell - 1} \mathbb{P}_m[w \mathrel{|} xy_{:i}] \right]\text{.}
    \end{align*}
    Thus, by \autoref{lemma} we have
    \begin{align*}
    &\mathrel{\phantom{=}} f_{\ell + 1}(w \mathrel{|} x) \\
    &= f_{\ell}(w \mathrel{|} x) + \mathbb{E}_{|y| = \ell}[\mathbb{P}_m[w \mathrel{|} xy]] \\
    &= \mathbb{E}_{|y| = \ell}\left[\mathbb{P}_m[w \mathrel{|} xy] + \sum_{i = 0}^{\ell - 1} \mathbb{P}_m[w \mathrel{|} xy_{:i}]\right] \\
    &= \mathbb{E}_{|y| = \ell}\left[\sum_{i = 0}^{\ell} \mathbb{P}_m[w \mathrel{|} xy_{:i}]\right]
    \end{align*}
    as desired.
\end{proof}

\section{Additional Results}
\label{sec:more-figs}
\subsection{Optimal values of $\beta$ and $\gamma$}
\begin{figure*}[h]
    \centering
    {\includegraphics[width=1\textwidth]{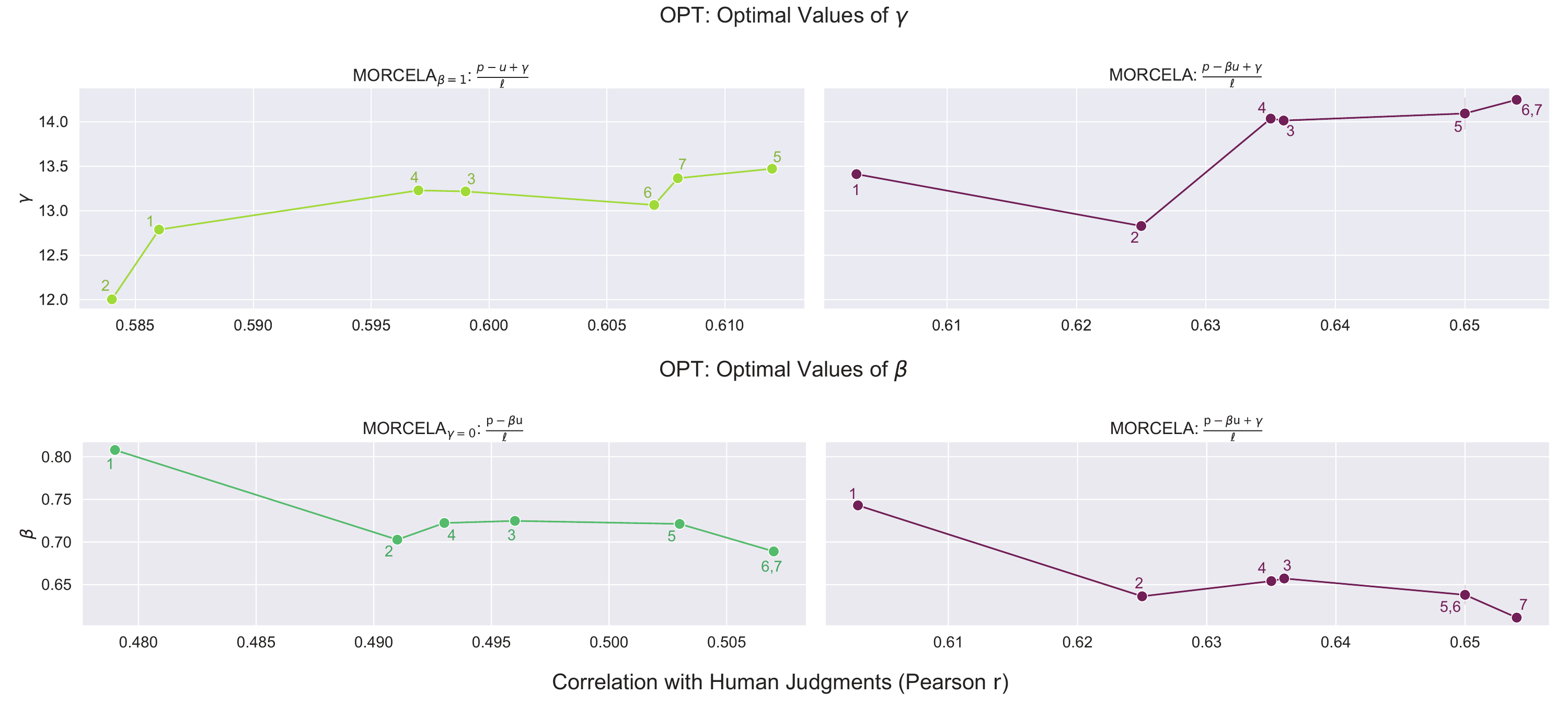}}
    {\includegraphics[width=1\textwidth]{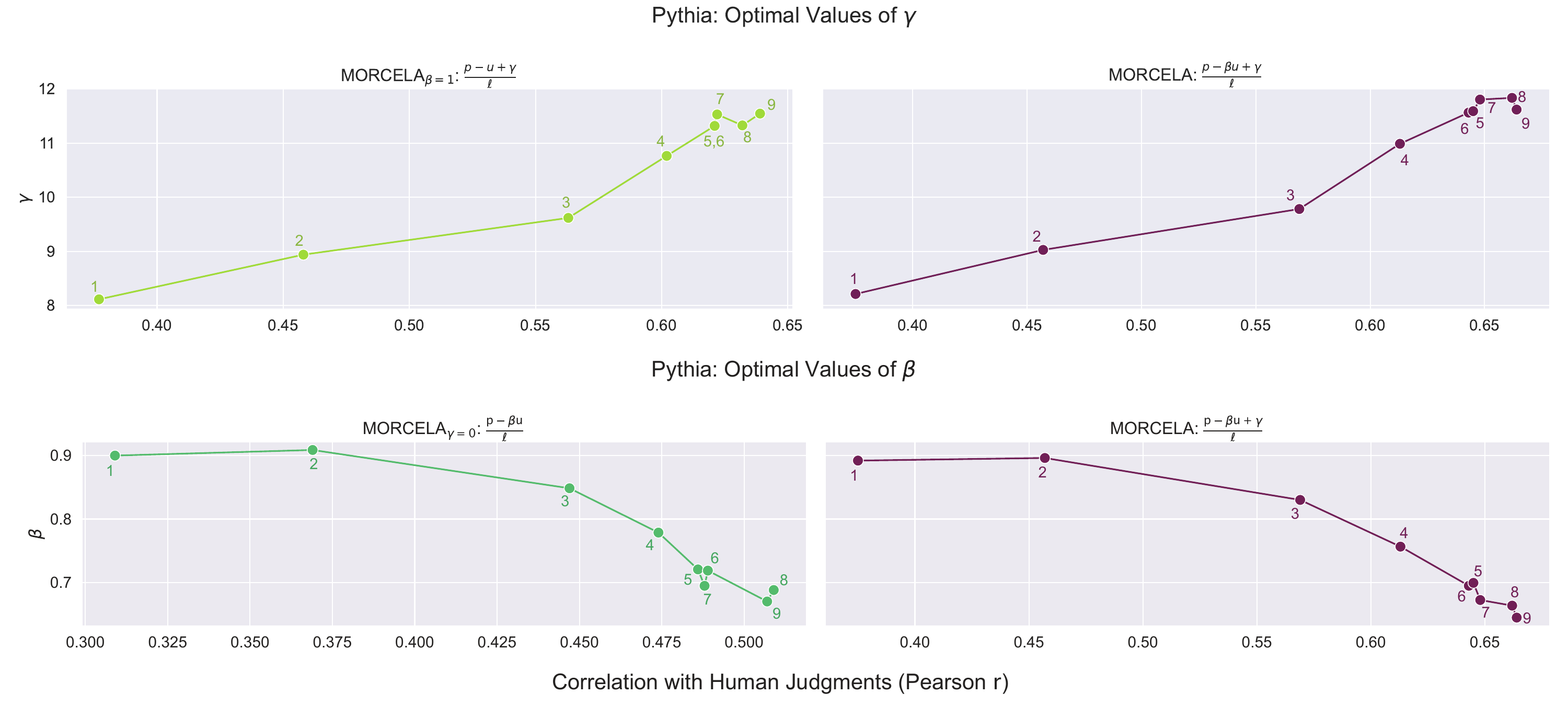}}
    \caption{Optimal values of $\gamma$ and $\beta$ versus correlation with human judgments across linking theories. Top set of plots are for the OPT models, bottom are for Pythia. Points are numbered in order of model size (smallest to largest).}%
    \label{fig:all-params}
\end{figure*}

Optimal values of MORCELA and the two ablations (MORCELA$_{\beta = 1}$ and MORCELA$_{\gamma = 0}$) for OPT and Pythia models are shown in \autoref{tab:beta-gamma} and visualized in \autoref{fig:all-params}.

\subsection{AIC and BIC Calculations}
\label{sec:aic_bic}
We calculate AIC and BIC for SLOR, MORCELA$_{\beta = 1}$, MORCELA$_{\gamma = 0}$, and MORCELA using the following formulas, where SSE is the sum of squared error and $n$ and $k$ are the sample size and number of predictor terms (including the intercept), respectively:
\begin{equation}
\text{AIC} = n * \ln{\frac{\text{SSE}}{n}} + 2k
\end{equation}
\begin{equation}
\text{BIC} = n * \ln{\frac{\text{SSE}}{n}} + k * \ln{n}
\end{equation}

AIC and BIC for each linking function for every model is reported in \autoref{tab:opt_aic_bic} and \autoref{tab:pythia_aic_bic}. Rows in each table are sorted in ascending order using BIC. For both AIC and BIC, a lower value is better.

\begin{table}[ht!]
    \centering
    \small
    \begin{tabular}{lccc} \toprule
    Model Variant & $\beta$ & $\gamma$ & $r$ \\ \midrule
    OPT 125M & 0.743 & 13.410 & 0.606 \\
    OPT 350M & 0.640 & 12.827 & 0.751 \\
    OPT 1.3B & 0.657 & 14.103 & 0.637 \\
    OPT 2.7B & 0.654 & 14.034 & 0.637 \\
    OPT 6.7B & 0.649 & 14.267 & 0.648 \\
    OPT 13B & 0.627 & 13.916 & 0.652 \\
    OPT 30B & 0.611 & 14.246 & 0.653 \\ \midrule
    Pythia 14M & 0.892 & 8.211 & 0.375 \\
    Pythia 70M & 0.896 & 9.026 & 0.457 \\
    Pythia 160M & 0.830 & 9.782 & 0.569 \\ 
    Pythia 410M & 0.756 & 10.987 & 0.613 \\
    Pythia 1B & 0.699 & 11.590 & 0.645 \\
    Pythia 1.4B & 0.695 & 11.563 & 0.643 \\
    Pythia 2.8B & 0.672 & 11.805 & 0.648 \\
    Pythia 6.9B & 0.664 & 11.863 & 0.662 \\
    Pythia 12B & 0.645 & 11.619 & 0.664 \\ \bottomrule
    \end{tabular}
    \caption{Optimal values of $\beta$ and $\gamma$ for each model, along with their correlation (Pearson $r$) with human judgments.}
    \label{tab:beta-gamma}
\end{table}
\begin{table}[h]
\small
    \centering
    \resizebox{\columnwidth}{!}{
    \begin{tabular}{llcccc}
    \toprule
    Size & Linking Function & AIC & BIC & SSE & Predictors \\
    \midrule
    \multirow{4}{*}{125M} 
    & MORCELA & -1384.3 & -1363.2 & 555.1 & 4 \\
    & MORCELA$_{\beta = 1}$ & -1329.5 & -1313.7 & 577.2 & 3 \\
    & MORCELA$_{\gamma = 0}$ & -1101.2 & -1085.4 & 675.7 & 3 \\
    & SLOR & -1091.2 & -1080.7 & 681.3 &  2 \\
    \midrule
    \multirow{4}{*}{350M} 
    & MORCELA & -1448.8 & -1427.7 & 530.9 & 4 \\
    & MORCELA$_{\beta = 1}$ & -1323.5 & -1307.7 & 579.6 & 3 \\
    & MORCELA$_{\gamma = 0}$ & -1133.5 & -1117.6 & 660.8 & 3 \\
    & SLOR & -1101.8 & -1091.3 & 676.3 & 2 \\
    \midrule
    \multirow{4}{*}{1.3B}
    & MORCELA & -1472.9 & -1451.8 & 522.2 & 4 \\
    & MORCELA$_{\beta = 1}$ & -1363.6 & -1347.8 & 563.8 & 3 \\
    & MORCELA$_{\gamma = 0}$ & -1128.8 & -1112.9 & 663.0 & 3 \\
    & SLOR & -1105.6 & -1095.0 & 674.6 & 2 \\
    \midrule
    \multirow{4}{*}{2.7B}
    & MORCELA & -1472.6 & -1451.4 & 522.3 & 4 \\
    & MORCELA$_{\beta = 1}$ & -1360.3 & -1344.5 & 565.1 & 3 \\
    & MORCELA$_{\gamma = 0}$ & -1126.0 & -1110.2 & 664.2 & 3 \\
    & SLOR & -1102.3 & -1091.7 & 676.1 & 2 \\
    \midrule
    \multirow{4}{*}{6.7B}
    & MORCELA & -1520.8 & -1499.7 & 505.2 & 4 \\
    & MORCELA$_{\beta = 1}$ & -1401.8 & -1386.0 & 549.2 & 3 \\
    & MORCELA$_{\gamma = 0}$ & -1150.7 & -1134.9 & 653.0 & 3 \\
    & SLOR & -1126.0 & -1115.4 & 665.2 & 2 \\
    \midrule
    \multirow{4}{*}{13B}
    & MORCELA & -1524.5 & -1503.4 & 503.9 & 4 \\
    & MORCELA$_{\beta = 1}$ & -1386.4 & -1370.6 & 555.0 & 3 \\
    & MORCELA$_{\gamma = 0}$ & -1155.3 & -1139.4 & 651.0 & 3 \\
    & SLOR & -1124.3 & -1113.7 & 665.9 & 2 \\
    \midrule
    \multirow{4}{*}{30B}
    & MORCELA & -1535.6 & -1514.4 & 500.1 & 4 \\
    & MORCELA$_{\beta = 1}$ & -1389.3 & -1373.4 & 553.9 & 3 \\
    & MORCELA$_{\gamma = 0}$ & -1156.7 & -1140.9 & 650.3 & 3 \\
    & SLOR & -1124.0 & -1113.4 & 666.1 & 2 \\
    \bottomrule
    \end{tabular}
    }
    \caption{
    AIC and BIC of various linking functions across OPT models.}
    \label{tab:opt_aic_bic}
\end{table}
\begin{table}%[h]
\small
    \centering
    \resizebox{\columnwidth}{!}{
    \begin{tabular}{llcccc}
    \toprule
    Size & Linking Function & AIC & BIC & SSE & Predictors \\
    \midrule
    \multirow{4}{*}{14M} 
    & MORCELA$_{\beta = 1}$ & -934.8 & -919.0 & 757.9 & 3 \\
    & MORCELA & -935.6 & -914.5 & 756.4 & 4 \\
    & SLOR & -782.2 & -771.6 & 843.1 & 2 \\
    & MORCELA$_{\gamma = 0}$ & -780.7 & -764.8 & 842.8 & 3 \\
    \midrule
    \multirow{4}{*}{70M} 
    & MORCELA & -1150.0 & -1128.9 & 652.4 & 4 \\
    & MORCELA$_{\beta = 1}$ & -1141.2 & -1125.3 & 657.3 & 3 \\
    & SLOR & -974.3 & -963.8 & 738.5 & 2 \\
    & MORCELA$_{\gamma = 0}$ & -972.7 & -956.9 & 738.3 & 3 \\
    \midrule
    \multirow{4}{*}{160M} 
    & MORCELA & -1341.4 & -1320.3 & 571.7 & 4 \\
    & MORCELA$_{\beta = 1}$ & -1307.7 & -1291.8 & 586.0 & 3 \\
    & SLOR & -1044.9 & -1034.3 & 703.4 & 2 \\
    & MORCELA$_{\gamma = 0}$ & -1045.6 &-1029.7 &  702.1 & 3 \\
    \midrule
    \multirow{4}{*}{410M} 
    & MORCELA & -1464.9 & -1443.8 & 525.1 & 4 \\
    & MORCELA$_{\beta = 1}$ & -1395.6 & -1379.8 & 551.5 & 3 \\
    & MORCELA$_{\gamma = 0}$ & -1105.2 & -1089.3& 673.8 & 3 \\
    & SLOR & -1099.5 & -1089.0 & 677.4 & 2 \\
    \midrule
    \multirow{4}{*}{1B} 
    & MORCELA & -1454.0 & -1432.9 & 529.0 & 4 \\
    & MORCELA$_{\beta = 1}$ & -1352.5 & -1336.7 & 568.2 & 3 \\
    & MORCELA$_{\gamma = 0}$ & -1090.2 & -1074.3  & 680.9 & 3 \\
    & SLOR & -1078.0 & -1067.4 &  687.5 & 2 \\
    \midrule
    \multirow{4}{*}{1.4B} 
    & MORCELA & -1446.0 & -1424.9 & 531.9 & 4 \\
    & MORCELA$_{\beta = 1}$ & -1353.0 & -1337.1 & 568.0 & 3 \\
    & MORCELA$_{\gamma = 0}$ & -1082.8 & -1067.0 & 684.3 & 3 \\
    & SLOR & -1072.6 & -1062.1 & 690.1 & 2 \\
    \midrule
    \multirow{4}{*}{2.8B} 
    & MORCELA & -1527.3 & -1506.1 & 503.0 & 4 \\
    & MORCELA$_{\beta = 1}$ & -1417.1 & -1401.3 & 543.4 & 3 \\
    & MORCELA$_{\gamma = 0}$ & -1113.4 & -1097.6 & 670.0 & 3 \\
    & SLOR & -1101.8 & -1091.2 & 676.3 & 2 \\
    \midrule
    \multirow{4}{*}{6.9B} 
    & MORCELA & -1560.6 & -1539.5 & 491.5 & 4 \\
    & MORCELA$_{\beta = 1}$ & -1451.9 & -1436.1 & 530.5 & 3 \\
    & MORCELA$_{\gamma = 0}$ & -1128.0 & -1112.1 & 663.3 & 3 \\
    & SLOR & -1117.4 & -1106.8 & 669.1 & 2 \\
    \midrule
    \multirow{4}{*}{12B} 
    & MORCELA & -1539.3 & -1518.1 & 498.8 & 4 \\
    & MORCELA$_{\beta = 1}$ & -1413.2 & -1397.3 & 544.9 & 3 \\
    & MORCELA$_{\gamma = 0}$ & -1113.5 & -1097.6 & 670.0  & 3 \\
    & SLOR & -1099.2 & -1088.6 & 677.6  & 2 \\
    \bottomrule
    \end{tabular}
    }
    \caption{
    AIC and BIC of various linking functions across Pythia models.}
    \label{tab:pythia_aic_bic}
\end{table}

\end{document}